# Efficient Medical Image Retrieval Using DenseNet and FAISS for BIRADS Classification

MD Shaikh Rahman [a*], Feiroz Humayara [b], Syed Maudud E Rabbi [c], Muhammad Mahbubur Rashid [d]

[a] Department of Computer Science, Universiti Sains Malaysia, Penang, Malaysia
[b] Department of Biomedicine, School of Dental Sciences, Universiti Sains Malaysia, Kelantan, Malaysia
[c] Baruch College, The City University of New York, New York, USA
[d] Faculty of Mechatronics Engineering, International Islamic University Malaysia, Kuala Lumpur, Malaysia

**Abstract**

That datasets that are used in today's research are especially vast in the medical field. Different types of medical images such as X-rays, MRI, CT scan etc. take up large amounts of space. This volume of data introduces challenges like accessing and retrieving specific images due to the size of the database. An efficient image retrieval system is essential as the database continues to grow to save time and resources. In this paper, we propose an approach to medical image retrieval using DenseNet for feature extraction and use FAISS for similarity search. DenseNet is well-suited for feature extraction in complex medical images and FAISS enables efficient handling of high-dimensional data in large-scale datasets. Unlike existing methods focused solely on classification accuracy, our method prioritizes both retrieval speed and diagnostic relevance, addressing a critical gap in real-time case comparison for radiologists. We applied the classification of breast cancer images using the BIRADS system. We utilized DenseNet's powerful feature representation and FAISS's efficient indexing capabilities to achieve high precision and recall in retrieving relevant images for diagnosis. We experimented on a dataset of 2006 images from the Categorized Digital Database for Low Energy and Subtracted Contrast Enhanced Spectral Mammography (CDD-CESM) images available on The Cancer Imaging Archive (TCIA). Our method outperforms conventional retrieval techniques, achieving a precision of 80% at k=5 for BIRADS classification. The dataset includes annotated CESM images and medical reports, providing a comprehensive foundation for our research.

## 1 Introduction

Breast cancer is a major health concern especially for women. To improve survivability and reduce mortality rates, ensuring early detection and accurate diagnosis is very important. Approximately 2.3 million new cases of breast cancer which leading to nearly 685,000 deaths globally in the year 2020 according to ("Breast Cancer Risk Genes — Association Analysis in More than 113,000 Women," 2021). Several techniques have been developed to for the purpose of detecting breast cancer. Mammography is one of the most effective tools to identify abnormalities in breast tissue which might cause cancer ("Mammogram Procedure | Johns Hopkins Medicine,"

n.d.). Breast Imaging Reporting and Data System also known as BIRADS is created by the American College of Radiology to standardize the evaluation of mammograms (Spak et al., 2017). This system provides a framework that categorize breast findings. These categories have a range of 0 to 6 where higher value indicate a greater chance of malignancy.

Analysis of medical images is very different compared to even few years ago. Traditional image retrieval methods relied on handcrafted features like texture, shape, and intensity to represent image content. However, these methods often fall short when dealing with the complexity of medical images. Deep learning has changed this landscape by allowing the automatic extraction of more complex features through CNNs, leading to improved performance in content-based image retrieval (CBIR) tasks. Specially since the integration of Deep learning methods. With deep learning, we are able to automate many complex tasks that earlier required expert interpretation. Different deep learning techniques have been used in the medical field. Among these, Convolutional Neural Networks or CNN in short-have remarkably advanced the field of medical image analysis (Alzubaidi et al., 2021). The big advantage of CNN is that it can directly learn from images hierarchical features, ranging from simple edges in the first layers to complex patterns deeper in the layers. Because of this capability in learning and identifying patterns within images, CNNs have widely been utilized (Yamashita et al., 2018), leading to state-of-the-art results in various diagnostic tasks.

Deep architecture of CNNs for breast cancer detection lets them capture fine-grained details relevant for diagnosis. Some of the challenges are related to limited data, class imbalance, and the need for fine-grained classification. (Anwar et al., 2023). Many CNN architectures, including ResNet (Al-Haija and Adebanjo, 2020), VGG (Liu et al., 2024), and Inception (Al Husaini et al., 2021), have been applied to breast cancer detection with promising outcomes. DenseNet stands out because of its unique design. DenseNet uses dense connections between layers, which helps in feature propagation and improves the reuse of learned features (Zhou et al., 2022). This makes it suitable for medical imaging tasks where the tiniest detail is important in the images for diagnosis. Even though CNN works very fine in image classification, retrieving similar images due to learned features is still a challenging task. In particular, it gets tougher for large medical image databases. The effectiveness of image retrieval provides fruitful insights into follow-up and abnormalities, thereby helping decision-making in medical practice.

However, even with advanced feature extraction, the efficient retrieval of images in large-scale medical databases remains a challenge. The nature of feature vectors extracted from deep networks often requires efficient indexing and search algorithms that guarantee fast retrieval. Such a problem could be dealt with using the open-source library Facebook AI Similarity Search, FAISS. FAISS is efficient for large-scale similarity searches. ("Faiss: A library for efficient similarity search - Engineering at Meta," n.d.). FAISS has supported various distance metrics such as L2 distance and cosine similarity, flexible for various image retrieval tasks. Large datasets search with FAISS is very efficient, a thing that was absolutely necessary for real-time medical imaging applications. FAISS is optimized for handling similarity searches in large-scale datasets and high-dimensional data using techniques such as product quantization and hierarchical indexing. The above-mentioned makes FAISS a strong tool to operate on medical image retrieval systems where rapid searches through extended datasets are allowed.

In this paper, we propose a method that effectively couples the strengths of DenseNet for robust feature extraction and FAISS for fast similarity search to retrieve relevant medical images for BIRADS classification. This will improve both the accuracy and speed of the image retrieval process to support clinical decisions. We test the effectiveness of this approach on a dataset of

2006 mammograms; it turns out to be effective with respect to precision at different depths of retrieval.

## 1.1 Objectives and Contributions

In this paper, we investigate how dense feature extraction using DenseNet combined with FAISS fared in the classification of mammograms. We will use the pre-trained DenseNet model for extracting feature vectors from mammographic images. These vectors will then be indexed by FAISS for the fast retrieval of similar images.

Key contributions of our work will include:

1. FAISS-DenseNet: We propose the architecture that extracts features with DenseNet and conducts image searching using. We want to facilitate retrieving mammograms that are similar to the query image for aiding in accurate BIRADS classification.
2. Extensive Evaluation: We evaluate the performance of our method on 2006 mammograms covering the broad range of BIRADS categories. We use different metrics like precision, recall, and normalized discounted cumulative gain to evaluate how the retrieval process is done.
3. Distance Metric Impact: We employ various distance metrics, including L2 distance and cosine similarity, to understand how they would impact the retrieval process and their suitability.
4. Visualization and Analysis: We employ both PCA and t-SNE visualizations to study the feature space learned by DenseNet. It shall provide insight into the distribution and clustering of different BIRADS categories in the embedding space.

It addresses the main challenges of medical image retrieval and thus proposes a practically feasible solution by combining deep learning and similarity search techniques to further develop automated diagnosis processes in mammography.

## 2 Related Work

For deep learning, similarity search methods, and feature extraction strategies, there has been extensive research on CBMIR and classification. Herein, we discuss the related works, especially those using deep neural networks and similarity search frameworks such as the DenseNet model and the FAISS library.

### 2.1 Convolutional Neural Networks (CNNs) for Medical Image Retrieval

CNNs have shown remarkable success in learning hierarchical feature representations from medical images (Thakur et al., 2024). Their multi-layer architecture allows them to capture low to high levels of features representative of essential characteristics like texture, edges, shapes, and patterns helpful to distinguish the medical conditions in the imaging data. While CNNs are poor at capturing sophisticated patterns inherent in medical images, due to automatically learning discriminative features directly from the imaging data, CNNs have formed the cornerstone for CBMIR systems in modern times. (Cai et al., 2019). Similarly, deep learning-based feature extraction has been explored for various medical imaging modalities, including mammography, ultrasound (Perdios et al., 2022), MRI (Farooq et al., 2017), and CT scans (Gupta et al., 2018). CNNs also had promising results on retrieving images among mammographic images while learning features of the breast tissue and helping in identifying abnormalities (Qayyum et al.,

2017). These works highlight that deep learning approaches can effectively handle the high intra-class variation and noise typical of medical imaging datasets.

## 2.2 Deep Hashing for Efficient Image Retrieval

Deep hashing techniques are able to transform high-dimensional image features into compact binary codes. This will enable fast retrieval while maintaining semantic similarity. A novel deep hashing-based framework, "Deep Triplet Hashing Network," was specifically designed for a case-based medical image retrieval network. This employed a triplet-based learning approach that encourages similar images-for example, images from patients having the same condition-to have similar hash codes and push dissimilar images apart in the hash space. (Fang et al., 2021). Their study demonstrated the effectiveness of deep hashing in achieving rapid retrieval with high precision, making it suitable for real-time CBMIR applications.

## 2.3 Transfer Learning and Fine-tuning of Pre-trained Models

The idea of transfer learning is very widely considered a current strategy in the community of CBMIR. Transfer learning makes use of pre-trained CNN models; most of them are ResNet, VGG, or DenseNet. Most of the CNN models applied in the transfer learning have already been trained on large general-purpose image databases such as ImageNet (Salehi et al., 2023). Such models can be further fine-tuned by researchers with medical imaging data, using the learned representations to adapt them to the peculiar characteristics of medical images. A study done in, which used a pre-trained CNN model fine-tuned on a dataset of mammographic images, has shown that transfer learning not only speeds up the process of model training but results in higher retrieval accuracy compared to training CNN from scratch on small medical datasets. (Guan and Loew, 2017).

## 2.4 Semi-Supervised Learning in Medical Imaging

Most of the existing SSL methods in medical imaging essentially focus on mining useful features from a large number of unlabeled images to guide model training (Huynh et al., 2022a). The key idea is to harness these features in the learning process itself for the representation, which later can be utilized for various downstream tasks such as classification, retrieval, and segmentation. Fairly importantly, this is utilized in medical image retrieval studies where the performance of retrieval significantly depends on the model capability for extracting rich and discriminative features.

Feature-oriented SSL methods employ a mixture of supervised learning on labeled data with either unsupervised or self-supervised methods on unlabeled data. In the proposed methodology, consistency loss is replaced by the ABCL approach within the SSL semi-supervised learning process (Huynh et al., 2022b). The alternative would be that the supervised model was trained by the labeled dataset in learning BIRADS-specific features, where the large amount of unlabeled data was incorporated using SSL methods that guide feature extraction (Zhang et al., 2020). The proposed framework showed that the inclusion of unlabeled data in the learning process enhanced the power of the model in distinguishing different BIRADS categories, thus improving the retrieval accuracy of the CBMIR. However, this is enhanced with the introduction of distance correlation via addition of different views of the same image and deep neural architectures provide better performance relative to many state-of-the-art semi-supervised learning (Berenguer et al., 2024). The application of feature-oriented SSL methods has demonstrated that the combination of labeled and unlabeled data presents more generalizable and robust feature representations. This is especially crucial in medical image retrieval, since medical images quite often exhibit high intra-class variability and subtle inter-class differences. In such cases, leveraging unlabeled images can help SSL-based models learn a wide range of visual features, thus improving retrieval performances for diverse cases. Despite all the promising advances of feature-oriented SSL in medical image analysis, some problems still persist: first, most of the time, the success of SSL methods will be consistently dependent on the quality of the pseudo-labels or the consistency of the features derived from the unlabeled data. The noisy pseudo-labels or inconsistent feature representations may introduce biases and affect the general performance of the model. Specifically, these limitations drive the researchers to put more emphasis on robust self-supervised learning methods such as contrastive learning that are designed to learn more general features with less dependence on pseudo-labels.

### *2.5 DenseNet for Medical Imaging*

These feature-reusing, deep-connected Densely Connected Convolutional Network architectures have thus found extensive applications in medical imaging. Jha et al. proposed a. The study epitomizes the efficiency and adaptiveness of DenseNet while handling such complex datasets of medical images, whereby a multichannel architecture for breast cancer lesion classification was sought to validate the versatility of this model in processing multi-view mammographic data and further establish the application of the proposed concept in CBMIR tasks. (Pawar et al., 2022).

The main reason for popularity in medical image retrieval is that DenseNet has dense connectivity, allowing feature re-use, which alleviates the problem of the vanishing gradient compared to deep networks. Besides, each layer in DenseNet is such that the input from all previous layers is fed into it, due to which much richer feature maps are obtained with enhanced learning of complicated patterns from the medical images. A 4-layer dense block is shown for DenseNet in Figure 1. The idea here is that each layer in the network shall be feed-forward connected to all previous layers, allowing it to reuse features and drastically improve gradient flow. This philosophy of connectivity further aids in developing a more detailed pattern recognition mechanism important in the interpretation of complicated medical images. The DenseNet has been proved to extract robust features in the light of retrieval accuracy in various imaging modalities. While CNN-based techniques have kept medical image retrieval a notch higher, there is still more work that needs to be done in order to curb some of the challenges. First, the complexity and variability within medical images require large-scale annotated data for effective training of deep models. It is extremely time-consuming and expert-intensive to annotate medical images, which results in limited availability of annotated data. Researchers are hence focusing on semi-supervised learning and self-supervised techniques that could leverage unlabeled data and improve feature learning (Zhang et al., 2020).

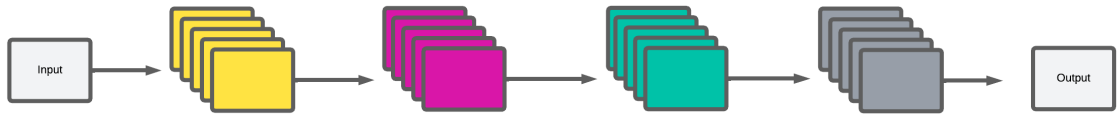

**Figure 1:** A 4-layer dense block in DenseNet architecture. Each layer receives inputs from all preceding layers, promoting feature reuse and efficient gradient flow, which results in improved learning of complex patterns.

## *2.6 FAISS for Similarity Search*

The FAISS library had grown to be an emerging tool for efficient similarity searches in large-scale datasets. Their main application areas are ANN searches, short for Approximate Nearest Neighbor. It allows for fast search queries in high-dimensional vector spaces. The FAISS library is designed to optimize the trade-off between the speed of the search and its accuracy.

(Douze et al., 2024). It employs several indexing methods, such as Inverted File Index-IVF and Product Quantization-PQ, to handle large-scale datasets effectively. These indexing methods enable FAISS to support both exact and approximate search depending on the application requirements. Full support of GPU and distributed computing in FAISS enhances its ability to handle trillions of data points and hence makes it ideal for industries dependent on large-scale image and text databases.

FAISS also combines well with cloud-based FPGA for remarkable speed and energy gains (Danopoulos et al., 2019). In this work, we show that FAISS, by integrating a variety of computing platforms, meets high computation demands when dealing with big data and thus is very suitable for cloud deployments.

The other application area for FAISS is in mass spectral library searches, whereby it integrates with FPGA-based accelerators (Zhang et al., 2018). This customized system uses indexing in FAISS to manage the complex and high-dimensional data from mass spectrometry; hence, the operation is optimized for high-speed performance while sacrificing none of the accuracy so critical in real-time scientific data analytics.

The fact that FAISS can be used for text retrieval or in image recommendation systems enhances its adaptability. For instance, in recommendation systems, the goal of FAISS might be to retrieve efficient embeddings that represent either users or items. This flexibility makes FAISS broadly applicable in everything from media content to text-based applications.

## 3. Methods

The section below describes our dataset, the preprocessing steps, feature extraction using DenseNet121, indexing using FAISS that aids in efficient image retrieval, and the evaluation metrics that were used to test the performance of our model.

### 3.1 Dataset

The data used for this paper are the Categorized Digital Database for Low Energy and Subtracted Contrast Enhanced Spectral Mammography (CDD-CESM) publicly available via The Cancer Imaging Archive (TCIA) *(Khaled et al., 2022)*. It consists of CESM images in high resolution with their annotation and medical reports. The low-energy and subtracted images were converted from DICOM to JPEG format at an average resolution of 2355 x 1315 pixels. Medical reports are in DOCX, while segmentation annotations are in CSV format, describing features such as mass shape, margins, density, and BIRADS assessment. The images were acquired using a dual-energy imaging protocol, where two exposures (low energy and high energy) were taken and recombined to suppress background breast parenchyma. Each examination took approximately 5-6 minutes, capturing both craniocaudal (CC) and mediolateral oblique (MLO) views for each subject. Figure 2 illustrates example images from the dataset: images (a) and (b) show low-energy images with minimal or no processing, while images (c) and (d) are subtracted images, digitally processed to enhance points of interest.

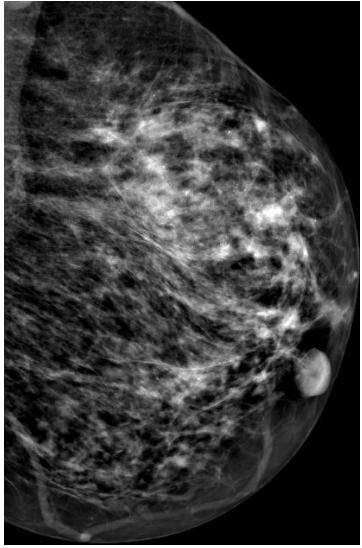 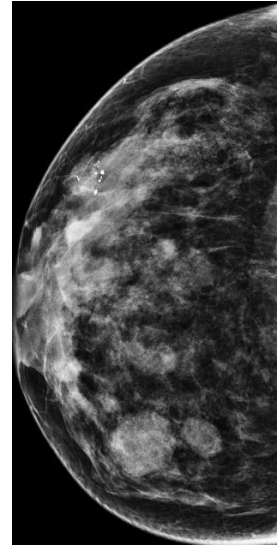

(a)                          (b)

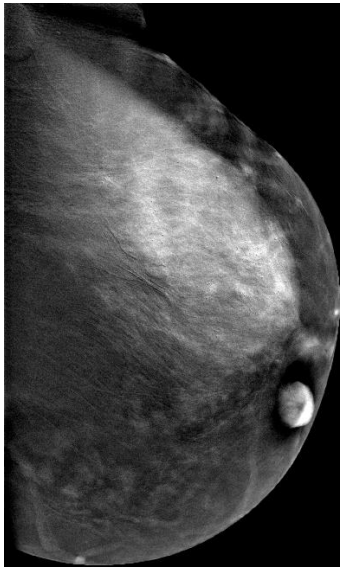 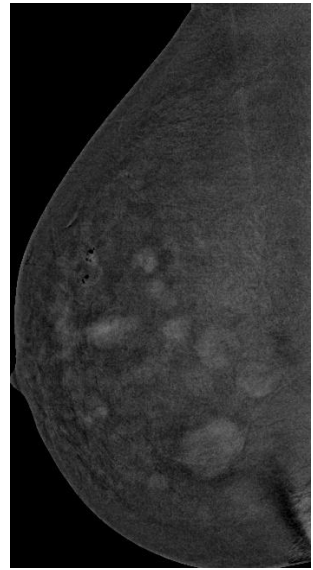

(c)                          (d)

**Figure 2:** Sample images from the CDD-CESM dataset. Images (a) and (b) represent the Low-Energy Images, which are the mammographic scans that have the least processing applied. Images (c) and (d) are Subtracted Images that have been digitally enhanced to outline regions of interest for better and easier identification of abnormalities

For this study, we used a dataset of 2006 images, classified using the Breast Imaging Reporting and Data System (BIRADS).

The dataset is organized into two categories:

- Resized Low-Energy Images: 1003 images that are low-energy representations of mammographic scans.

- Resized Subtracted Images: 1003 images that have been digitally processed to highlight areas of interest in the scans.

Each image is linked to a BIRADS score, which ranges from 1 to 6, where:

- BIRADS 1: Negative (no findings),

- BIRADS 2: Benign findings,

- BIRADS 3: Probably benign,

- BIRADS 4: Suspicious abnormality,

- BIRADS 5: Highly suggestive of malignancy,

- BIRADS 6: Known biopsy-proven malignancy.

The BIRADS labels were provided in an Excel sheet, which contained the image filename, BIRADS score, and other metadata such as patient ID and findings. The data was evenly distributed across the different BIRADS categories, providing a balanced set of images for testing the retrieval system.

### 3.2 Feature Extraction

We employed DenseNet121, a deep convolutional neural network pre-trained on ImageNet, for feature extraction due to its efficient parameter use and ability to extract high-quality image features. The classifier layer was removed from DenseNet121 to enable deep feature embedding extraction. The feature extraction process involved the following steps:

1. Image Preprocessing:

    - The images were resized to 224x224 pixels, matching the input size expected by DenseNet121.

    - Each image was converted from grayscale to a 3-channel format to match the structure of the DenseNet input.

    - Normalization was applied using the mean and standard deviation of the ImageNet dataset, which DenseNet was pre-trained on.

2. Feature Embedding:

    - The processed images were passed through the modified DenseNet121 (without the classifier) to generate a feature vector for each image.

    - The output of DenseNet121 was a 1024-dimensional feature vector, which served as the image's high-level representation in the retrieval process.

### 3.3 FAISS Indexing

For fast and accurate image retrieval, we utilized **FAISS** (Facebook AI Similarity Search), a library optimized for similarity search and clustering of dense vectors. Two types of distance metrics were explored:

1. L2 Distance (Euclidean): Measures the straight-line distance between two feature vectors.

2. Cosine Similarity: Measures the cosine of the angle between two vectors, often used when the magnitude of the vectors is less important than their orientation.

The extracted feature vectors from DenseNet121 were indexed using FAISS, and two indexing methods were used depending on the distance metric:

- **FlatL2 Indexing** for Euclidean distance: A flat (brute-force) index that stores the feature vectors and performs linear searches based on L2 distance.

- **FlatIP Indexing** for cosine similarity: This uses inner product searches to approximate cosine similarity, with L2 normalization applied to the feature vectors to convert inner product calculations into cosine similarity.

After building the FAISS index, we could perform efficient similarity searches for any given query image by retrieving the top-k nearest neighbors from the index.

## 3.4 Evaluation Metrics

The performance of the retrieval system was evaluated using the following metrics:

1. Precision at k: Measures the proportion of retrieved images (top-k) that share the same BIRADS score as the query image. Precision at k is crucial for evaluating the accuracy of the retrieval system (Olson and Delen, 2008), particularly in medical imaging where false positives can mislead diagnosis.

$$Precision\ at\ k = \frac{Number\ of\ relevant\ images\ in\ top-k}{k} \quad (1)$$

For instance, a precision at k = 5 of 0.80 indicates that 4 out of the 5 retrieved images have the same BIRADS score as the query image.

2. **Recall at k**: Measures the proportion of relevant images that have been retrieved in the top-k results out of all relevant images in the dataset. This metric is significant when assessing whether the system can retrieve all possible relevant images for a given query (Powers and Ailab, n.d.).

$$Recall = \frac{True\ Positives}{True\ Positives + False\ Negatives} \tag{2}$$

3. **Normalized Discounted Cumulative Gain (NDCG)**: Measures the ranking quality of the retrieved results. NDCG considers both the relevance and the position of each image in the ranked list. Higher relevance scores are weighted more heavily if they appear earlier in the retrieval list (Wang et al., 2013), which is crucial in a medical context where accurate images should be prioritized.

$$NDCG\ at\ k = \frac{1}{IDCG_k} + \sum_{i=1}^{k} \left( \frac{2^{rel_{\{i\}}} - 1}{\log_{\{2\}}(i+1)} \right) \tag{3}$$

Where $rel_i$ is the relevance of the image at position **i** in the top-k results. $IDCG_k$ is the **Ideal Discounted Cumulative Gain** at position k, which is the maximum possible DCG for the top k ranked items.

4. **Search Time**: The time taken to retrieve the top-k results for a given query. Search time is important in real-time applications where the system must return results quickly.

5. **Visualization**: We used principal component analysis (PCA**)** (Wang et al., 2013) and t-distributed Stochastic Neighbor Embedding (t-SNE) (Maaten and Hinton, 2008)for visualization. Both are dimensional reduction methods which helps in exploring dataset. PCA was applied to reduce the feature space while retaining the maximum variance in the dataset, helping us visualize clusters of similar images in lower dimensions. t-SNE was employed to capture the local structure of the data and reveal patterns and relationships among images that might not be apparent in higher dimensions. Both techniques provide insights into how the deep learning model organizes and distinguishes between different BIRADS categories in the feature space.

These metrics were computed for different values of **k** (k = 1, 5, 10, 20, 50) to analyze how the system performs with varying numbers of retrieved images. In this study, the primary objective is to evaluate the retrieval performance of multiple deep learning models using FAISS for image similarity search. Unlike traditional machine learning approaches that require dataset splitting (training, validation, test sets) to assess model generalization, our focus is on feature extraction and retrieval accuracy within a fixed dataset. Therefore, no dataset splitting was performed, as this work emphasizes performance metrics related to image retrieval rather than classification accuracy or model generalization

### *3.5 Comparison with Other Models*

To validate the effectiveness of the DenseNet121 + FAISS combination, we also compared its performance with other well-known architectures like **ResNet50** and **VGG16**. Similar experiments were conducted with these models, using the same evaluation metrics (precision, recall, NDCG) to compare their performance in the BIRADS classification task. The comparison focused on retrieval accuracy, search speed, and feature embedding quality.

The results show that DenseNet121 combined with FAISS provides a good balance between retrieval accuracy and efficiency, making it suitable for use in medical image retrieval applications.

## 4    Results

### 4.1  Optimization Process and Experimental Setup

First step of our experiment consists of running Densenet121 using Faiss for image search. To find the right parameters for optimized retrieval we ran the code with multiple values of k (the number of nearest neighbors retrieved): 1, 5, 10, 20, 50 and 100. This allowed us to assess how retrieval accuracy differs with different number of neighbors. To measure the key performance, we used metrics such as precision, recall and NDCG.

### 4.2  Results for DenseNet121

For DenseNet121, the retrieval performance at different **k** values is summarized below:

- Precision: DenseNet121 achieved a precision of **0.8** at **k=5**, indicating that 80% of the top 5 retrieved images were relevant. This increases to **0.9** when **k=10**. When the value of k is increased further the precision slowly decreases but still keeps significant precision of 0.8 or higher. Precision still maintains 0.8 when k=50 and becomes 0.79 when k=100 We then change the image and tried with 10 more images and the precision values almost stays constant from 0.8 to 0.9 at k=10. From table 1 we can clearly observe the change of precision for different image and k.

**Table 1:** Precision, Recall, NDCG, and Search Time for DenseNet121 at Various Values of k. The table presents retrieval performance metrics for DenseNet121, showing precision, recall, normalized discounted cumulative gain (NDCG), and search time (in seconds) for different values of k and query images. As k increases, precision initially improves, reaching 0.9 at k=10, but slightly decreases at higher values, stabilizing around 0.8. NDCG values indicate high ranking quality, particularly for smaller values of k, while search times remain negligible, indicating the efficiency of the retrieval system.

| Query Image | k | Precision | Recall | NDCG | Search Time Sec |
|---|---|---|---|---|---|
| 0 | 1 | 1 | 0.001248 | 1 | 0.00104 |
| 0 | 5 | 0.8 | 0.004994 | 0.982892 | 0.001086 |
| 0 | 10 | 0.9 | 0.011236 | 0.966715 | 0 |
| 1 | 1 | 1 | 0.001248 | 1 | 0 |
| 1 | 5 | 0.4 | 0.002497 | 1 | 0.001033 |
| 1 | 10 | 0.7 | 0.008739 | 0.886759 | 0 |
| 2 | 1 | 1 | 0.001248 | 1 | 0 |
| 2 | 5 | 1 | 0.006242 | 1 | 0.001048 |
| 2 | 10 | 0.9 | 0.011236 | 1 | 0 |
| 3 | 1 | 1 | 0.001248 | 1 | 0.001068 |
| 3 | 5 | 0.8 | 0.004994 | 0.982892 | 0 |
| 3 | 10 | 0.9 | 0.011236 | 0.966715 | 0 |
| 4 | 1 | 1 | 0.001248 | 1 | 0 |
| 4 | 5 | 1 | 0.006242 | 1 | 0.000992 |
| 4 | 10 | 0.9 | 0.011236 | 0.997188 | 0 |

- **Recall**: As the value of **k** increased, the recall improved, reaching **0.0499** at **k=50** for query image 0. This indicates that more relevant images were retrieved as the search window widened. This is also noticeable when k is set to 50 or 100 where the recall is then 0.05 and 0.09.

- **NDCG**: The NDCG metric, which measures the quality of the ranking of the retrieved images, was highest at **k=5** with a score of **0.98**, indicating that relevant images were ranked near the top of the retrieval list. This value slightly decreases when the value of k is increased. When **k=10** the score is **0.96** which is still a remarkable. When k is increased more like 50 or 100 the score varies slightly as 0.95 and 0.94

- **Time:** For each query, the search time was recorded to evaluate how quickly the system could find images. For **k=1**, the search time was **1.04 milliseconds**. As k increased, there was a slight increase in time, but it remained negligible, with 1.086 milliseconds for k=5. FAISS is extremely fast, and in some cases, the search time was so small that it was rounded down to zero.

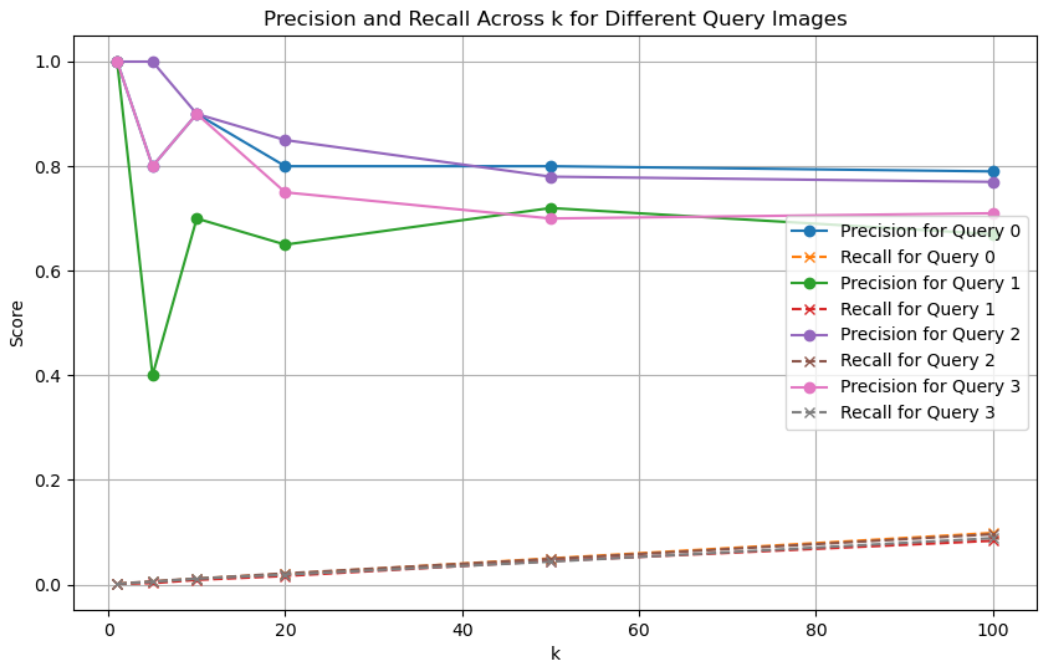

**Figure 3**: Precision and Recall scores across different values of k for various query images using FAISS indexing with DenseNet121. Precision generally maintains high levels for most queries at smaller values of k, while recall gradually increases with larger values of k, indicating improved retrieval of relevant images as the search space widens.

*4.3 Evaluation of Models Using Faiss Indexing*

To properly evaluate the performance of DenseNet121 with FAISS indexing, we also tested other models such as ResNet50, VGG16, and EfficientNet. Table 2 presents the comparative results across these models.

**ResNet50:** When FAISS indexing is applied using ResNet50, it demonstrates a precision comparable to DenseNet121. For k=5, ResNet50 achieves a precision of 0.8, which matches DenseNet121. However, precision drops significantly to 0.6 when k=10. Increasing the value of k slightly improves precision, reaching a maximum of 0.7 at k=20. Similar to precision, the recall for ResNet50 at k=5 is 0.005, close to that of DenseNet121, but it remains consistently lower than DenseNet121 as k increases. In terms of search time, ResNet50 is slightly slower than DenseNet121, but the NDCG values are comparable between the two models.

**VGG16:** VGG16 shows a noticeable decrease in performance across precision, recall, and NDCG compared to DenseNet121. For query image 0 at k=5, VGG16 achieves a precision of 0.4, which is significantly lower. The corresponding recall and NDCG values for VGG16 are 0.002 and 0.85, respectively. The performance fluctuates considerably across different images, leading to inconsistent precision, recall, and NDCG values. Additionally, the search time for VGG16 is higher than for DenseNet121, making it less efficient in retrieval tasks.

**EfficientNet:** While EfficientNet shows promising results for some queries, its performance is less stable compared to DenseNet121. For image 0, it delivers a precision of 0.8 and an NDCG of 0.96 at k=5, closely matching DenseNet121. However, when querying different images, the performance drops significantly. For instance, for image 1, the precision decreases to 0.4, and NDCG drops to 0.85. Although EfficientNet can perform well for certain images, its inconsistency across different queries affects its overall reliability.

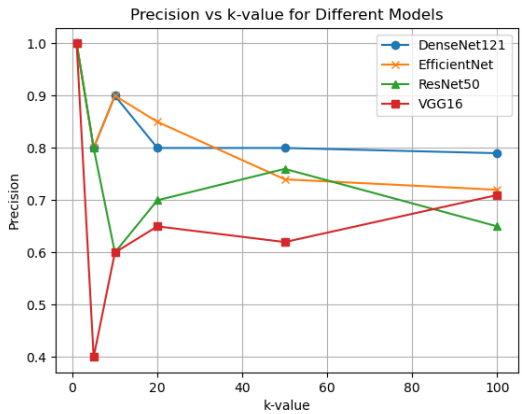

(a)

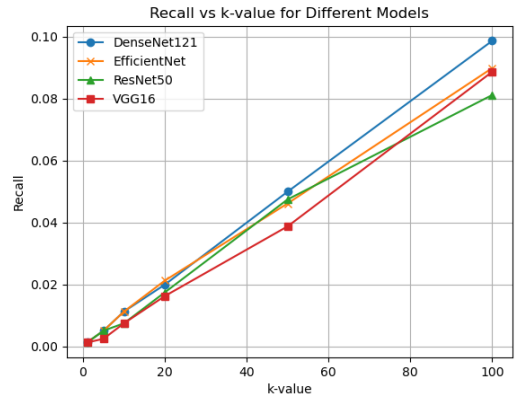

(b)

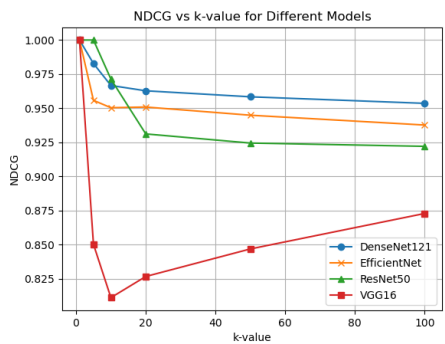

(c)

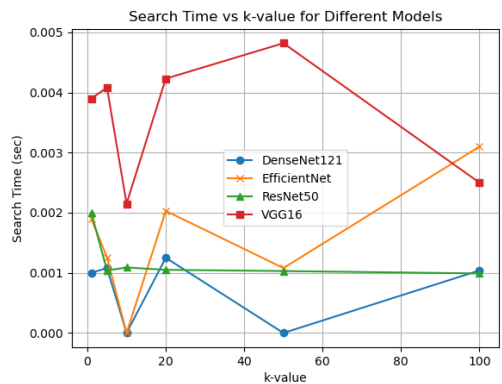

(d)

**Figure 4:** Comparison of retrieval performance metrics across different models (DenseNet121, EfficientNet, ResNet50, and VGG16) as k-values increase. Panel (a) illustrates precision vs. k, where DenseNet121 achieves the highest precision at 0.8, followed by EfficientNet and VGG16 at 0.7, with ResNet50 lagging below 0.65. Panel (b) shows recall vs. k, with recall consistently increasing across all models as k increases, and DenseNet121 achieving the best recall at higher k-values. In panel (c), NDCG vs. k is displayed, with DenseNet121 consistently demonstrating superior ranking quality, followed closely by EfficientNet, while VGG16 shows the lowest NDCG scores. Panel (d) presents search time vs. k, where DenseNet121 maintains the lowest search time, highlighting its efficiency, while VGG16 exhibits the highest search times.

From **Figure 4**, we can clearly observe the differences in performance between the four models. The first plot illustrates the **precision** for all models as **k** increases. DenseNet121 achieves a precision of **0.8**, while VGG16 and EfficientNet reach **0.7**, and ResNet50 lags behind with a precision below **0.65**.

Similarly, both the **recall** and **NDCG** metrics demonstrate that DenseNet121 outperforms the other three models. The final plot highlights the **search time**, where DenseNet121 consistently shows the fastest performance, having the lowest retrieval time compared to the other models.

### 4.4 Visualization of Embedding Space

We used **PCA** and **t-SNE** visualizations to better understand how DenseNet121 organizes images. For this we reduce the high-dimensional embeddings to two dimensions. These visualizations will help to illustrate the structure of the embeddings and how well they capture meaningful patterns from the dataset.

**PCA Visualization:** In Figure 5 (a), the PCA plot shows how the first two principal components explain 82% of the variance in the feature space. Each point in the plot represents an image, and its color corresponds to the BIRADS category. The clusters formed in the PCA plot demonstrate that DenseNet121 successfully captures some structure in the data, as points close to each other likely have similar embeddings, although the separation between BIRADS categories is subtle.

**t-SNE Visualization:** In Figure 5 (b), the t-SNE visualization of the embedding space (after PCA) provides an alternative view, focusing on the local structure of the data. Here, the images that have similar embeddings are grouped closer together. The plot shows a more defined separation between different images, further illustrating how DenseNet121 captures underlying relationships between images in the dataset.

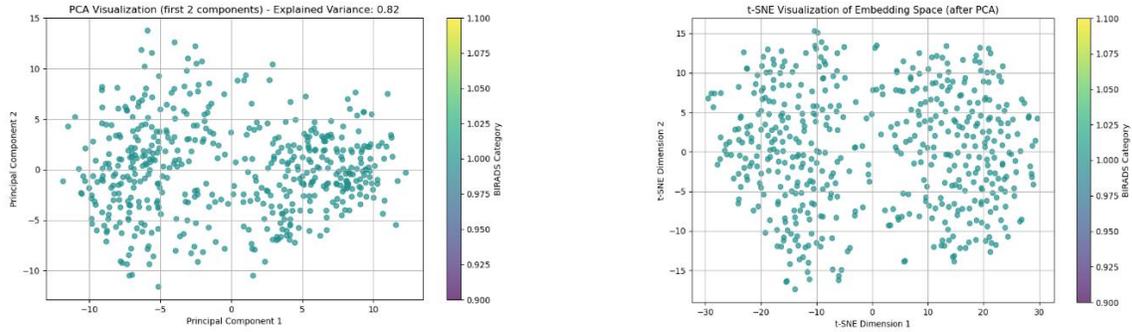

**Figure 5:** Visualization of DenseNet121 Embedding Space for BIRADS Categories. (a) PCA Visualization: The first two principal components explain 82% of the variance in the feature space. (b) t-SNE Visualization: This view focuses on local structure after PCA, grouping similar images more closely.

Both the PCA and t-SNE visualizations show that DenseNet121 effectively extracts features from the dataset, resulting in meaningful embeddings that can be used for image retrieval.

## 5  Discussion

The results we obtained from our experiment determines that **DenseNet121** used with FAISS indexing is able to perform well. In terms of precision, recall, and NDCG across multiple values of *k* our model showed good results. To gain a deeper understanding of its effectiveness, we analyze these findings in the broader context of image retrieval performance by comparing it with other models such as ResNet50, VGG16, and EfficientNet.

### 5.1 DenseNet121 Performance

Densenet121 showed consistent performance for the various ranges of k. It achieved a precision of 0.8 at k=5, with a maximum of 0.9 for k=10, still retaining a high precision value of 0.79 at k=50 and k=100. This would thus confirm its ability to retrieve relevant images even within large search spaces. This is important in real applications, since large databases of images are often the rule rather than the exception, and faster retrieval becomes increasingly crucial. Also, DenseNet121 yielded a high-ranking quality of the top-retrieved images, with an NDCG of 0.98 at k = 5. The small drop of NDCG for growing k, 0.94 at k = 100, suggests that while precision remains high, the relative order of retrieved images starts to deteriorate as the retrieval window opens further.

DenseNet has a unique architecture where each layer receives the feature maps from all preceding layers, enabling feature reuse and a densely connected structure. This design helps in capturing fine-grained details and learning richer, more diversified features that are beneficial for medical images, especially where minor variations are critical (e.g., distinguishing subtle tissue abnormalities).

Despite its strong retrieval performance, one limitation of DenseNet121 is its recall. As shown in the results, the recall remains low at 0.05 when $k=50$ and does not significantly improve with higher $k$ values. This suggests that while DenseNet121 excels at retrieving the most relevant images, it may not be capturing all relevant images in the search space. This trade-off between precision and recall is important to consider, especially in medical or diagnostic applications, where failing to retrieve relevant images could lead to missing important information.

### *5.2 Comparison with Other Models*

Among these compared models, the performance of DenseNet121 had the best precision, NDCG, and retrieval time, which makes it suitable for being applied to large datasets.

- **ResNet50:** ResNet50 was close to DenseNet121, but mostly at the lower values of k. For example, at k = 5, precision for ResNet50 is as high as 0.8 for DenseNet121, but it decreases rather fast when moving to higher k-for instance, a performance at k = 10 with precision = 0.6. Recall and NDCG of ResNet50 lag behind DenseNet121, meaning that it fails to maintain relevant rankings in a larger window. While powerful for small retrieval tasks, ResNet50 might not scale up as well as DenseNet121 for larger datasets.

- **VGG16:** VGG16 tended to give the worst performance among the compared models. Precision at k = 5 had fallen as low as 0.4, which is very low against both DenseNet121 and ResNet50. Besides, its performance metrics are fluctuating across queries and value changes of k in general, showing unstable and inconsistent results with regard to image retrieval tasks. Such instability might probably be due to the architecture of VGG16, perhaps not effective in extracting discriminating features compared to DenseNet121 or EfficientNet. Furthermore, VGG16 searches more slowly compared with VTR, so that VGG16 does not fit retrieval tasks on large-scale occasions.
- **EfficientNet:** The results of EfficientNet have shown promises in achieving high precision and NDCG at some queries, mainly for the small values of k. For example, at k=5, it achieved a precision of 0.8, very close to DenseNet121. It got an NDCG of 0.96, indicating the appropriateness of ranking in the retrieved images. However, the model has performed inconsistently across queries. For example, EfficientNet's precision dropped to 0.4 for k=5 with different query images, suggesting difficulties with generalization. This model seems more sensitive to the query image, which may limit its applicability in highly diverse datasets.

Though powerful, ResNet50 was designed to alleviate problems of vanishing gradient with its skip connections, rather than ensure dense connectivity. The overall architecture of VGG16 is very simple and deeper in a sequential manner, with no reuse of the features in between. This results in a limited capture of intricate multi-scale patterns within the images. Retrieval performance by VGG16, because of the simple architecture, fails to capture intricate patterns, which are essential in medical imaging. The presence of skip connections in ResNet50 makes it more general, hence may miss the needed specialty to catch minute details in the mammographic image. Therefore, the capability of DenseNet for feature propagation and reusing makes it highly suitable for medical image retrieval since it would learn features which are descriptive yet highly distinctive across various categories, like BIRADS classification. Its architecture also allows for a far more efficient flow of gradients. This leads to better convergence and stable learning even with deeper networks.

## 5.3 Search Efficiency

One of the notable strengths of DenseNet121 is its efficiency in image retrieval, particularly in terms of search time. As shown in the results, DenseNet121 consistently had the lowest search time compared to the other models. Even as the value of $k$ increased, the search time remained negligible, rarely exceeding 1 millisecond. In a field as important as the health sector, both the accuracy of retrieval and the speed at which results are returned play a critical role in user experience.

Models like VGG16 exhibited lower precision, recall and also took longer to perform searches, making them less efficient for real-time or large-scale retrieval tasks. EfficientNet, while performing well in terms of precision for some queries, also showed relatively higher search times, indicating that it may not be as computationally efficient as DenseNet121 for large datasets.

## 5.4 Visualizing the Embedding Space

We used two visualizations to get further insights of our model. PCA and t-SNE visualizations provide insight into how DenseNet121 organizes the images in the feature space.

The PCA plot explained 82% of the variance using the first two principal components, showing that DenseNet121 captures a significant amount of information in its embeddings. However, the clustering of BIRADS categories was not strongly defined, indicating that DenseNet121 does not fully separate different categories in the feature space, which may contribute to the lower recall observed in the retrieval tasks.

The **t-SNE** visualization, which focuses on the local structure of the embeddings, shows a clearer separation between images, with clusters forming more distinctly than in the PCA plot. This suggests that while DenseNet121 is able to capture local relationships well (as reflected by its high precision and NDCG), the global structure of the data may not be as effectively represented, leading to the observed trade-offs in recall.

### 5.5  Limitations and Future Work

Despite DenseNet121's strong performance, there are several areas where improvements can be made. The low recall at higher *k* values suggests that DenseNet121 might be missing some relevant images, which is particularly important in domains such as medical imaging, where it is crucial to retrieve all relevant cases. Additionally, while search time is generally low, it will be important to test the system on larger datasets to ensure that the performance remains consistent as the size of the search space grows. While this study focuses on evaluating the retrieval performance within a single dataset, future work could explore the use of dataset splitting or separate query and gallery sets to assess retrieval in more varied conditions. However, since the emphasis of this paper is on feature extraction and similarity-based image retrieval, the current approach without dataset splitting is adequate to address the primary research question

Future work could explore hybrid models that combine the strengths of multiple architectures or investigate more advanced indexing techniques to improve recall without sacrificing precision. Furthermore, fine-tuning DenseNet121 on specific image retrieval tasks could enhance its ability to capture domain-specific features, improving both precision and recall.

## 6    Conclusion

In this paper, we introduced a novel approach for medical image retrieval, focusing specifically on the classification of mammographic images using DenseNet121 for feature extraction and FAISS for efficient similarity search. Our strategy, in this work, has been the enhancement of speed and accuracy in image retrieval concerning the BIRADS classification, which is a very critical activity in the diagnosis of breast cancer.

In our experiment, DenseNet121 combined with FAISS indexing showed consistently high precision and NDCG scores for different k values, hence powerful in retrieving relevant images even in large search spaces. With this model, it retained the same degree of precision with an increase in the number of images retrieved. In other words, it scaled well without losing its precision within a large database. This was also with retrieval time reduced significantly compared to competing models. Because of this, DenseNet121 proved to be a great candidate for medical applications that required real-time responses.

Among these, the DenseNet121 performed better in both retrieval accuracy and efficiency compared to other models such as ResNet50, VGG16, and EfficientNet. ResNet50 was performing pretty well on the small-scale retrieval tasks but tumbled down and failed to catch up with the increasing retrieval window. VGG16 underperformed across all metrics, with fluctuating results and longer search times. While EfficientNet showed promising results for specific queries, its overall performance was inconsistent across various images, indicating challenges in generalization.

We also visualized the feature space learned by DenseNet121 using PCA and t-SNE, which provided insights into how the model organizes the images. The PCA plot explained 82% of the variance using the first two principal components, showing that DenseNet121 captures substantial information in its embeddings. However, the clustering of BIRADS categories was not strongly defined, suggesting that DenseNet121 does not completely separate different categories in the feature space, which may contribute to the lower recall observed in some retrieval tasks. The t-SNE visualization illustrated a clearer separation between images, indicating that DenseNet121 effectively captures local relationships, as evidenced by its high precision and NDCG scores.

While DenseNet121 shows considerable promise, there are areas for improvement. The low recall at higher k values suggests that the model might miss some relevant images, an important consideration in medical imaging where retrieving all relevant cases can be critical. Although the search time was low in this study, testing the system on even larger datasets could help verify that its performance remains consistent as the dataset size grows.

In conclusion, the combination of DenseNet121 with FAISS indexing presents a robust and scalable solution for medical image retrieval tasks, particularly in BIRADS classification. Its ability to achieve high precision, recall, and efficiency highlights its potential to support radiologists in diagnostic decision-making. Future work could involve further fine-tuning of DenseNet121, exploring additional indexing strategies, or applying this approach to larger and more varied datasets to enhance its generalizability and ensure it meets the demands of clinical practice at scale.

*CRediT authorship contribution statement*

**MD Shaikh Rahman:** Writing – original draft**,** Writing – review and editing, Methodology, Investigation, Formal Analysis, Visualization, Data curation, Conceptualization, Project administration. **Feiroz Humayara:** Writing – review & editing, Writing – original draft, Data curation, Conceptualization. **Syed Maudud E Rabbi:** Writing – review & editing, Software, Methodology. **Muhammad Mahbubur Rashid:** Writing – review & editing, Validation, Supervision, Resources, Project administration.

*Declaration of Generative AI and AI-assisted technologies in the writing process*

During the preparation of this work the author(s) used GPT-4 from OpenAI in order to improve readability and language. After using this tool/service, the author(s) reviewed and edited the content as needed and take(s) full responsibility for the content of the publication.

*Declaration of competing interest*

The authors declare that they have no known competing financial interests or personal relationships that could have appeared to influence the work reported in this paper.

*Funding*

This research did not receive any specific grant from funding agencies in the public, commercial, or not-for-profit sectors.

*Data availability*

Data will be made available on request.